\documentclass[sigconf]{acmart}

\AtBeginDocument{%
  \providecommand\BibTeX{{%
    \normalfont B\kern-0.5em{\scshape i\kern-0.25em b}\kern-0.8em\TeX}}}

\setcopyright{acmcopyright}
\copyrightyear{2018}
\acmYear{2018}
\acmDOI{XXXXXXX.XXXXXXX}

\acmConference[Conference acronym 'XX]{Make sure to enter the correct
  conference title from your rights confirmation emai}{June 03--05,
  2018}{Woodstock, NY}
\acmBooktitle{Woodstock '18: ACM Symposium on Neural Gaze Detection,
 June 03--05, 2018, Woodstock, NY} 
\acmPrice{15.00}
\acmISBN{978-1-4503-XXXX-X/18/06}

\usepackage{subcaption}

\usepackage{rotating}
\usepackage{multirow}
\usepackage{array}
\usepackage{makecell}

\usepackage{glossaries}

\begin{document}

\title{Don't freeze: Finetune encoders for better Self-Supervised HAR}



\begin{abstract}
  Recently self-supervised learning has been proposed in the field of human activity recognition as a solution to the labelled data availability problem. The idea being that by using pretext tasks such as reconstruction or contrastive predictive coding, useful representations can be learned that then can be used for classification. Those approaches follow the pretrain, freeze and fine-tune procedure. In this paper we will show how a simple change - not freezing the representation - leads to substantial performance gains across pretext tasks. The improvement was found in all four investigated datasets and across all four pretext tasks and is inversely proportional to amount of labelled data. Moreover the effect is present whether the pretext task is carried on the Capture24 dataset or directly in unlabelled data of the target dataset. Code and data are available at \textit{git link}.
\end{abstract}




\received{20 February 2007}
\received[revised]{12 March 2009}
\received[accepted]{5 June 2009}

\author{Vitor Fortes Rey, Dominique Nshimyimana, Paul Lukowicz}
 \affiliation{
	\institution{DFKI, RPTU}
	\country{Germany}
 }
\email{fortes@dfki.uni-kl.de, nshimyim@rptu.de, Paul.Lukowicz@dfki.de}


\begin{teaserfigure}
  \centering
  \includegraphics[width=1.0\linewidth]{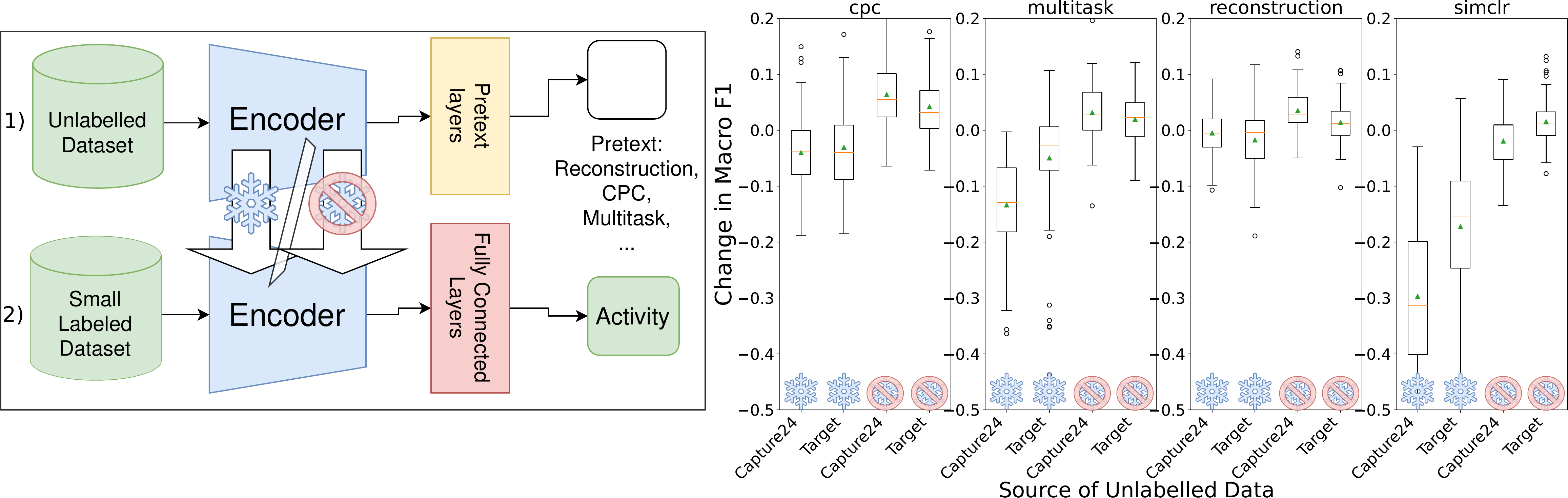}
  \caption{Change in performance given approach for using self-supervised learned representations in different SSL methods over four HAR datasets. }
  \label{fig:ssl}
  \Description{}
\end{teaserfigure}

\maketitle

\section{Introduction and Related Work}
Human Activity Recognition (HAR) with wearable sensors is an important application in the pervasive computing field. With the prevalence of inertial measurement units (IMUs) in devices such as smartwatches and smartphones, performing HAR using, for example, acceleration, is a very active field of research with decades of research.  
Although much has been achieved, the field still struggles with the lack of large annotated datasets. While obtaining IMU data is easy, annotating said data is not straightforward, as even experts cannot provide high quality labels based on, for example, just the accelerometer data. In fact, when collecting HAR data, labelling is often done using synchronised video from one or more cameras. This data collection and annotation is expensive and time consuming, and thus many works focus on advancing the state of the art despite the lack of data. Of note are two approaches: simulating IMU data from videos found in online repositories and self-supervised learning (SSL). While the simulation approach has found some success\cite{kwon2020imutube,fortes2021translating}, it suffers from problems beyond limitations in the simulation itself. First, it constrains activities to those that can be easily filmed. Second, human labelling is still necessary, which is not ideal.
On the other hand SSL relies on unlabelled data, simulated or real, to learn a useful representation that then can be fine-tuned for HAR in the target dataset. Learning this representation is done by performing a pretext task such as reconstructing the original signal\cite{reconstruction}, predicting which transformation was performed on the data\cite{multitask}, among others. Recently, the field has been moving to this framework of representation learning \cite{assessing_har,enhanced_cpc, multtask_oxford}.

Those approaches follow the pretrain, freeze and fine-tune procedure. In this paper we will show how a simple change - not freezing the encoder representation during fine-tuning - leads to substantial performance gains across pretext tasks. 
In other words in this paper we analyse four pretexts in four HAR datasets and compare:
\begin{itemize}
    \item the standard approach of fine-tuning with a frozen encoder versus fine-tuning the encoder representation and classification layers together, as seen in Figure \ref{fig:ssl}.
    \item learning representations in a bigger unlabelled dataset (Capture24) versus in the target dataset.
    \item how much the amount of labelled data affects the first two factors. 
\end{itemize}
Our findings can inform the development of new self-supervised approaches for HAR as well as guide the use of current ones, as our proposed fine-tuning showed improvement in all four investigated datasets and across all studied pretext tasks with gains inversely proportional to amount of labelled data. Moreover the effect is present whether the pretext task is carried on the Capture24 dataset or directly in unlabelled data of the target dataset.

For our tests we selected the pretexts as Multi-task \cite{multitask}, cpc\cite{cpc, enhanced_cpc}, Reconstruction \cite{reconstruction}, SimClr \cite{simclr}. Multi-task\cite{multitask, selfhar} proposes auxiliary tasks of binary classification for extracting useful features for the downstream task. A multi-task temporal convolutional neural network (CNN) is trained to classify whether the input signal was transformed or not for each of a set of transformations such as scaling, rotating the signal, etc. The authors utilize a large unlabelled accelerometer dataset for self-supervised learning and transfer learned knowledge to a small, labelled activity recognition dataset. cpc\cite{cpc,enhanced_cpc} uses an encoder that generates a representation for each timestep. Then, a specific point of time is selected and future step representations are predicted from the selected one using contrastive loss, where for each future timestep the positive sample is the next one in time and negative samples are representations for that time in other instances of the batch. Reconstruction\cite{reconstruction} is the application of auto-encoders where an encoder network reduces the dimensionality of the data while a decoder one brings the representation back to its original dimensions, with the whole system learning to reconstruct the original signal. SimClr \cite{simclr, contrastive_learning_har} learns general representations from a dataset where a model is trained to match different views instances created from the same signal by contrasting them. The positive pair is obtained by the transformation of the input window, while the negative pair is formed by the rest input windows of the same batch. Then, the contrastive loss pulls the positive pair together while pulling the negative pair far away from each other in the representation space.

Closest to our work is \cite{assessing_har} where SSL was systematically evaluated in the HAR domain. They studied the robustness to differing source and target conditions of SSL and the influence of dataset characteristics along with feature space ones, but always performed fine-tuning with a frozen encoder. More recents works \cite{multtask_oxford} also found improvement in not freezing the encoder, but they used for pre-training a dataset with over 7000 subjects that is not publicly accessible and analysed only the Multi-task\cite{multitask} pretext.
\section{Evaluation Procedure}
In this paper we evaluate several SSL methods under different training conditions. Our experiment considers only \textbf{accelerometer data} of a single IMU. Generally data are collected from sensors worn at diverse body positions. For example myogym\cite{myogym}, mhealth \cite{mHealth} and Capture24 \cite{capture24} datasets are Wrist worn while Motionsense \cite{motionsense} refers to Waist (left or right trousers pocket). Capture24 \cite{capture24} is a large-scale dataset collected in free living conditions from more than 150 users and is used for pre-training due to its large size. We use mHealth \cite{mHealth}, motionsense \cite{motionsense}, myogym \cite{myogym} and pamap2 \cite{pamap2} for evaluation of the HAR task. For all tests we under-sampled the data to 50Hz and used sliding windows of two seconds (100) with of one second (50) step size.For Capture24, we randomly select $16$ users for validation and use the remaining ones for training. For each target dataset we separate users randomly into 5 groups and perform cross validation so that in each fold 3 groups are used for training, one for testing and the remaining one for validation. The folds are done so that all groups are selected once for test and once for validation.
Each fold test is repeated 5 times with different random seeds, in order to obtain a better idea of the overall performance.

The first step of our pipeline is model selection for each pretext task. We performed the hyperparameter search for each method following the hyperparameters from \cite{enhanced_cpc} sampling $20$ variations. We trained each one for $50$ epochs, evaluating on the validation set
every epoch with a patience of 5. Our encoders are as in \cite{enhanced_cpc}, but we have selected $256$ as the size of our encoder representation (output), so for fairness we add a linear layer with relu activation to encoders from \cite{enhanced_cpc} that have a smaller representation size.
As this search was performed in Capture24, a large dataset,  we independently developed a similar sampling approach to \cite{multtask_oxford}. Every window has a sampling probability proportional to the sum of variance of each acceleration channel. At each epoch we sample 10\% of the full dataset. As in \cite{multtask_oxford}, we found this sampling approach to be beneficial to not only the computation time needed to train models, but also to their overall performance. This can be due to the fact that windows with actual movement convey more information, but it is also related to the fact that this sampling method approximates better the distribution of variance between Capture24 and the target datasets.
For pretexts that involve data augmentation with transformantions, such as Multi-Task and SimClr, one can either do the transformations before the training starts or compute new ones each epoch. We tested both approaches for our $20$ hyperparameters, and found that Multi-task has better results with the precomputed transformations. For fairness, other methods are also tested one more time on a fixed training set.
In all cases the best model for each pretext is determined based on its loss in the validation set of Capture24.
For training pretexts in the target dataset there are $5$ training folds each with a different training set so for each fold we always trained pretext tasks on that fold's \textbf{full} unlabelled training data and used for evaluation its validation set. 
In order to save computational resources, we have directly used the winning hyperparameters obtained in the Capture24 search.

In each test, the selected encoder model for each pretext is then used for HAR (see Figure \ref{fig:ssl}). We use as baseline its architecture without any pretext and compare that with training with the frozen learned encoder and training without freezing said encoder. Regarding classification we have used as classifier a network with $3$ layers of size $[256,128,128]$, each followed by batch norm, a relu activation and 0.2 dropout. Although a simple classifier with a linear layer provides the same overall conclusion regarding results (at lower performances), we omit those from the paper for brevity. Here we perform the same hyperparameter search as in \cite{enhanced_cpc}, but for every run also go over all hyperparameters with the SGD optimiser and no scheduler. The best model for each of the 5 runs is selected based on the best mean validation macro f1 across all folds of that run.

We also investigate the effect of data availability by reducing the amount of labels present for fine-tuning. We tested with 2,5,10,50 and 100 windows per class for all methods, but using always the full validation set. This test included the full hyperparameter search performed in the normal classification test. 

The exact user splits for each dataset, hyperparameters and details are accessible at \footnote{link to git repository}.
\section{Results and Discussion}
\begin{figure*}
    \centering
    \includegraphics[width=0.65\textwidth]{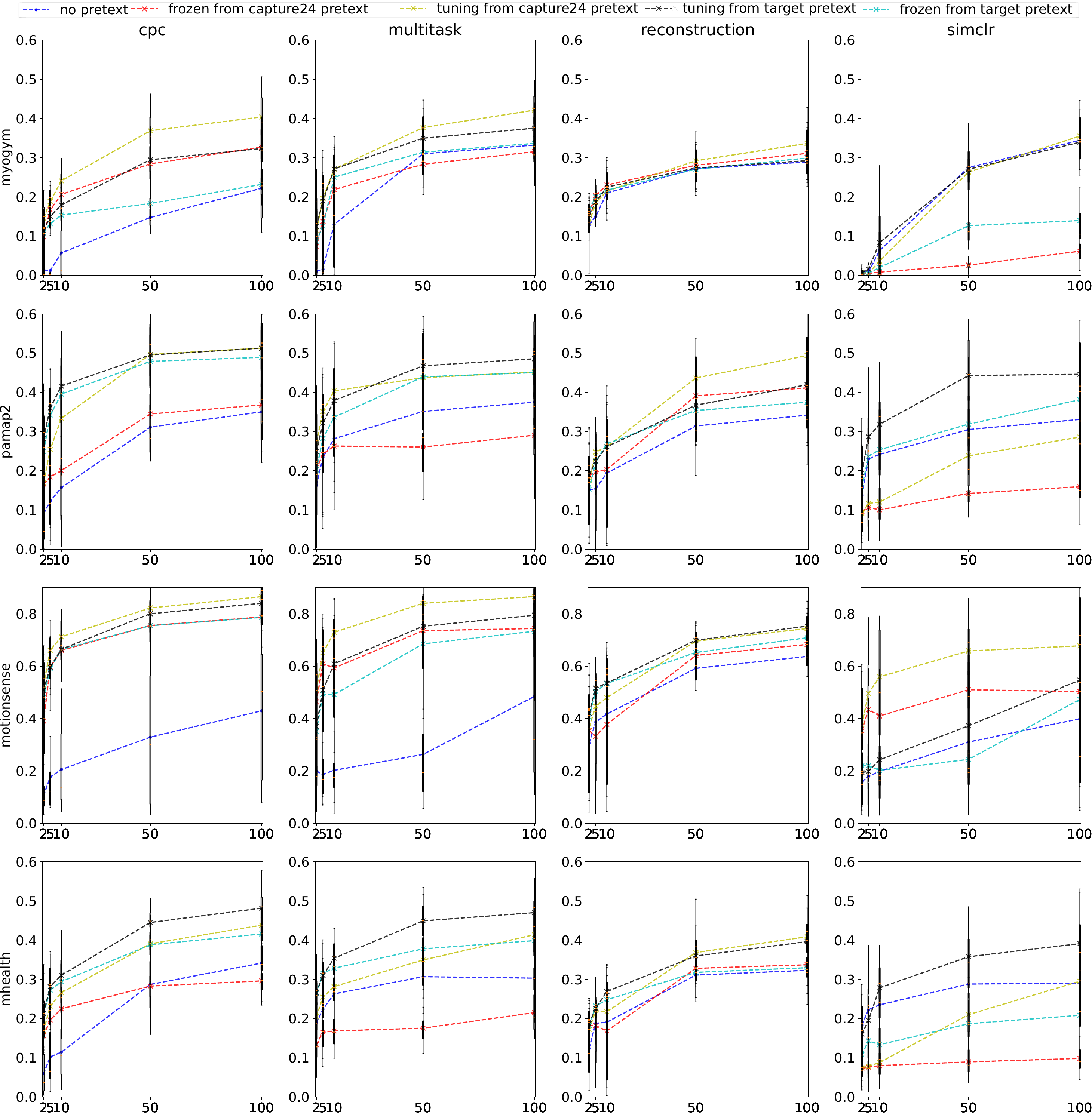}
    \caption{Performance of different training strategies when varying the amount of sampled labelled windows per class. The blue line represents learning only classification starting from scratch while other lines represent starting the encoder with pretext weights using different strategies regarding freezing or not the encoder and the dataset used for SSL.}
    \label{fig:data_red}
\end{figure*}

In Table \ref{table:mlp} we can see results for different pretexts over different datasets. Compared to the baseline results we can see that using frozen representations helps in some cases, but those improvements are overshadowed by using not frozen ones (either on Capture24 or on the target dataset). When fine-tuning the encoder classification performance improves across all datasets for most pretextes, with the best results being with the cpc and Multi-task methods, even if their baseline results are not the best for that dataset. The exception is the SimClr pretext, which decreases performance in the myogym and motionsense datasets when pretexts are done using Capture24 data\footnote{This is probably because augmentations were done using only rotation.}. In the case of motionsense this may be due to the change in body position, while for myogym the pretext representation does not seem to learn a useful representation in general, as can be observed in the performance of its frozen use. If we look at the distribution of change in macro F1 in relation to the baseline (no pretext), see Figure \ref{fig:ssl}, we can again see this overall trend of clearer improvement with full fine-tuning. Interestingly, using a bigger dataset (Capture24) for SSL provided better results, probably due to the hyperparamter search being performed over that dataset, but it is also possible that it being a bigger and more varied dataset makes up for the change in domain if the appropriate sampling is performed. 
It is also important to see the effect of labelled data availability, as we are interested in reducing labelled data requirements for HAR. From Figure \ref{fig:data_red} one can see that across datasets and pretexts we can often reach or surpass the performance of a baseline model with 50 or 100 windows per class by simply fine-tuning with 10 or fewer windows per class. This varies across pretexts and datasets, with cpc or Multi-task providing the most improvement.
\subsection{Pretext task overfitting}
Another relevant question is how to decide when to stop training the pretext model. In our previous results we stopped at the best validation epoch, but we have also explored training the model after that point. To evaluate this we took the best pretext models from Capture24, reinitialised their weights and performed the pretexts again at Capture24 followed by classification with the previosly selected hyperparameters and without freezing the encoder. This is done with representations at three points: before any pretext is done, at the point of best validation score and at the end of training (epoch 50). Results can be seen in Table \ref{table:pretext_self}, where it is clear that some methods, such as cpc, can overfit, while others, such as Multi-task and reconstruction, are less affected by the stopping point, with similar performance at the last epoch.
\begin{table}
\centering
\caption{Classification results (macro f1) when stopping the pretext at different points. SSL is performed in the Capture24 dataset and encoder not frozen when fine-tuning.}
\begin{tabular}{lllllll}
\toprule
Dataset                  & Pretext        & \makecell{Before \\ Pretext} & \makecell{Best \\ validation} & \makecell{Last \\ Epoch} \\
\midrule
 \multirow{4}{*}{ \makecell{motion-\\ sense} } 
                         & cpc            & 87.78±0.83 & \textbf{91.3±1.88}    & 89.14±1.0   \\
                         & simclr         & 85.17±2.24 & \textbf{85.22±2.65}   & 84.53±2.04 \\
                         & \small reconstruction & 87.14±0.71 & 88.04±0.7    & \textbf{88.11±0.8}  \\
                         & multitask      & 86.54±1.94 & 89.85±0.92   & \textbf{89.87±1.05} \\
                         \midrule
            pamap2       & cpc            & 53.87±1.14 & \textbf{57.39±4.82}   & 53.56±1.67 \\ 
                         & simclr         & \textbf{48.48±1.54} & 46.13±2.22   & 45.48±1.94 \\
                         & \small reconstruction & 52.01±1.54 & \textbf{52.65±1.99}   & 52.37±1.88 \\
                         & multitask      & 48.82±1.95 & \textbf{50.16±1.41}   & \textbf{50.16±1.81} \\
                         \midrule
            mhealth      & cpc            & 48.73±2.02 & \textbf{56.14±5.42}   & 52.88±2.19 \\  
                         & simclr         & \textbf{49.58±1.77} & 48.27±2.6    & 48.07±1.35 \\
                         & \small reconstruction & 46.54±1.62 & \textbf{48.72±1.95}   & 48.45±1.44 \\ 
                         & multitask      & 48.63±2.27 & \textbf{50.52±1.74}   & 50.49±1.77 \\
                         \midrule
            myogym       & cpc            & 44.32±4.74 & \textbf{55.43±5.89}   & 48.54±2.15 \\ 
                         & simclr         & \textbf{44.52±1.06} & 44.26±1.63   & 44.2±2.07  \\
                         & \small reconstruction & 45.78±1.52 & 47.92±0.95   & \textbf{48.01±1.31} \\ 
                         & multitask      & 44.7±1.28  & 51.54±1.39   & \textbf{51.72±0.81} \\

\bottomrule
\end{tabular}
\label{table:pretext_self}
\end{table}
\begin{table*}
\centering
\caption{Classification results (macro f1) results for different SSL methods over different datasets. }
\begin{tabular}{lllllll}
\toprule
    Dataset &        Pretext &  Baseline & Frozen Capture24 &   Frozen Target & Tuning Capture24 &  Tuning Target \\
\midrule
     myogym &            cpc & 39.37 ± 5.30 &   41.10 ± 5.46 &  29.44 ± 4.20 &   50.19 ± 6.57 & 44.41 ± 5.54 \\
      &         simclr & 46.15 ± 7.31 &    4.42 ± 0.88 &  16.96 ± 7.76 &   42.86 ± 6.48 & 47.47 ± 7.09 \\
      & reconstruction & 45.76 ± 5.85 &   42.04 ± 4.97 &  43.52 ± 7.02 &   47.74 ± 5.13 & 46.66 ± 6.26 \\
      &      multitask & 45.03 ± 6.67 &   37.81 ± 4.19 &  43.77 ± 4.66 &   \textbf{52.03 ± 5.40} & 49.46 ± 6.12 \\
     \midrule
     pamap2 &            cpc & 54.36 ± 7.74 &   46.18 ± 7.22 &  51.88 ± 8.90 &   \textbf{57.53 ± 7.29} & 56.50 ± 8.03 \\
      &         simclr & 49.68 ± 6.87 &   22.50 ± 6.19 &  45.16 ± 8.36 &   47.10 ± 8.02 & 52.62 ± 7.10 \\
      & reconstruction & 51.75 ± 6.34 &   51.07 ± 6.80 &  44.69 ± 8.51 &   56.53 ± 6.82 & 53.79 ± 7.04 \\
      &      multitask & 50.47 ± 8.05 &   35.38 ± 5.97 &  49.09 ± 7.03 &   52.13 ± 7.27 & 51.82 ± 8.77 \\
     \midrule
motionsense &            cpc & 86.96 ± 6.00 &   85.13 ± 4.16 &  82.91 ± 5.97 &   \textbf{91.05} ± 4.57 & 87.88 ± 6.20 \\
 &         simclr & 87.67 ± 4.37 &   75.84 ± 6.51 &  73.72 ± 6.60 &   85.53 ± 4.40 & 87.86 ± 4.45 \\
 & reconstruction & 85.59 ± 6.11 &   85.05 ± 4.72 &  86.56 ± 4.46 &   88.07 ± 4.69 & 86.93 ± 5.03 \\
 &      multitask & 86.97 ± 5.58 &   78.63 ± 3.60 & 77.30 ± 16.65 &   89.75 ± 3.51 & 88.57 ± 5.65 \\
\midrule
    mhealth &            cpc & 46.76 ± 7.60 &   39.03 ± 3.09 &  50.98 ± 2.51 &   54.23 ± 5.87 & \textbf{55.66 ± 6.65} \\
     &         simclr & 52.95 ± 5.91 &   14.93 ± 4.05 &  31.71 ± 5.53 &   53.16 ± 6.40 & 54.55 ± 6.02 \\
     & reconstruction & 48.24 ± 5.22 &   51.22 ± 4.59 &  49.61 ± 5.57 &   53.07 ± 4.52 & 49.47 ± 5.40 \\
     &      multitask & 53.38 ± 6.76 &   30.61 ± 6.08 &  45.89 ± 6.12 &   54.56 ± 5.12 & 53.83 ± 5.97 \\
\bottomrule
\end{tabular}
\label{table:mlp}
\end{table*}

\section{Conclusion and Future Work}
In this paper we have shown how SSL can be beneficial for HAR using unlabelled data either from the same dataset or from another one with enough variability. Following our suggested training protocols we can also reduce the need for labelled data, alleviating one of the major difficulties in developing HAR systems.
Still there are many open questions such as the effects of combining different datasets when performing SSL, or continuing the pretext task while training the overall classification layers. A more varied hyperparameter search for encoder and classification networks could also be done across pretexts. Even more fundamental, a deeper study of the relationship between current SSL approaches and the physical constraints of the target activities would be helpful in designing more informantive pretexts for HAR.

\clearpage
\bibliographystyle{ACM-Reference-Format}
\bibliography{sample-base}

\end{document}